\documentclass[lettersize,journal]{IEEEtran}
\usepackage{amsmath,amsfonts}
\usepackage{algorithmic}
\usepackage{algorithm}
\usepackage{array}
\usepackage[caption=false,font=normalsize,labelfont=sf,textfont=sf]{subfig}
\usepackage{textcomp}
\usepackage{stfloats}
\usepackage{url}
\usepackage{verbatim}
\usepackage{graphicx}
\usepackage{cite}
\usepackage{amsmath,amssymb}

\usepackage{gensymb}

\usepackage{adjustbox}

\usepackage{comment}

\begin{document}

%
\title{False Memory Formation in Continual Learners Through Imperceptible Backdoor Trigger}
%
%
%

\author{Muhammad~Umer
        and~Robi Polikar
\thanks{M. Umer and R. Polikar are with the Department
of Electrical and Computer Engineering, Rowan University, Glassboro,
NJ, 08028 USA.}
\thanks{Manuscript received ; revised .}}

\markboth{Journal of --- ,~Vol.~, No.~, Month~20--}%
{Umer \MakeLowercase{\textit{et al.}}:False Memory Formation in Artificial Neural Networks}
%



\maketitle

\begin{abstract}
In this brief, we show that sequentially learning new information presented to a continual (incremental) learning model introduces new security risks: an intelligent adversary can introduce small amount of \textit{misinformation} to the model during training to cause deliberate forgetting of a specific task or class at test time, thus creating ``false memory'' about that task. We demonstrate such an adversary's ability to assume control of the model by injecting ``backdoor'' attack samples to commonly used generative replay and regularization based continual learning approaches using continual learning benchmark variants of MNIST, as well as the more challenging SVHN and CIFAR 10 datasets. Perhaps most damaging, we show this vulnerability to be very acute and exceptionally effective: the backdoor pattern in our attack model can be imperceptible to human eye, can be provided at any point in time, can be added into the training data of even a single possibly unrelated task and can be achieved with as few as just \textit{1\%} of total training dataset of a single task.

\end{abstract}


%
\IEEEpeerreviewmaketitle

\section{Introduction}
\IEEEPARstart{H}{uman} memory is well-known to be susceptible to \textit{false memory formation} \cite{doi:10.1080/09658211.2019.1611862}, where one's memory can be easily distorted through post-event misinformation. To create such false memory, a malicious entity can provide deliberate and persistent misinformation over a period of time to convince an otherwise unsuspecting victim of the adversary's preferred -- but inaccurate -- version of events. In this effort, we show that a deliberate adversary can also cause false memory formation in artificial neural networks (ANNs) in continual/incremental learning of a sequence of tasks. While we certainly do not ascribe psychological or biological properties to ANNs, it is instructive to note the parallels between deliberate exposure to misleading information that distorts the memory of its victim vs. the memory of an ANN.

In this work, we pose the following question: can an adversary plant false memory (misinformation) into an ANN model in an incremental learning setting -- and in fact take advantage of the incremental learning setup -- in order to distort the model's memory while learning new tasks? We show that the answer is yes: such misinformation can be easily incorporated into the memory of a continually-learning ANN through adversarial \textit{backdoor} poisoning attacks. We also describe an attack model that can achieve its objective by inserting only a very small amount of misinformation into the training data of only a single task, where the misinformation can be as subtle as to be completely imperceptible to humans.
We demonstrate that using such a small and imperceptible attack, the adversary can force a model to learn the misinformation with great precision to have any class to be labeled with its intended false-label during test time, hence demonstrating the acuteness, severity and pinpoint effectiveness of such a poisoning attack against continually-learning ANNs. 

\section{Continual / Incremental Learning}
Continual learning (CL)---also referred to as life-long, sequential, or incremental learning---is the problem of learning tasks sequentially from a stream of data \cite{de2019continual}. A common phenomenon plaguing CL is \textit{catastrophic forgetting} \cite{mccloskey1989catastrophic}, where the ability of the model to recall a previously-learned task (e.g., a class) is partially or completely lost (forgotten) while the model is training to acquire new knowledge for learning a subsequent task.  Catastrophic forgetting is often characterized by the \textit{stability - plasticity dilemma} \cite{grossberg1988nonlinear}, where stability refers to ability of the model to preserve past knowledge, and plasticity is its ability to acquire and integrate new knowledge. 
There are three important CL scenarios described in the literature \cite{vandeven2019three}, depending on whether task identity is provided at inference time and, if not, whether it must be inferred: 
In \textit{task-incremental learning} the task identity is always provided both during training and inference time, resulting in a simple but also an impractical CL scenario as the model always knows which task needs to be trained or tested. 
In \textit{domain-incremental learning (DIL)} the task identity is not available at test time; however, the model also does not need to infer the task identity either. This scenario assumes that the structure of the task remains the same, i.e., the number of classes and class distribution do not change, but the input distribution is changing.  
In \textit{class-incremental learning (CIL)} the models must learn each task seen thus far \textit{and} infer with which task they are presented. CIL is the most challenging but also the most realistic scenario as it incrementally learns new classes over time. 

In this work, we consider the latter two more practical (and challenging) scenarios, and demonstrate that both are indeed vulnerable to small and imperceptible misinformation provided during training of even a single task of adversary's choice. With this misinformation, the adversary can then cause deliberate forgetting of a particular task at test time. 

\subsection{Regularization and Replay-based Continual Learning}
Regularization and replay-based approaches are two well known approaches in continual learning literature \cite{Buzzega2020DarkEF,de2019continual,Umer2020TargetedFA}.
\textit{Regularization-based approaches} add a regularization term to the loss function to prevent the loss of prior knowledge while learning new tasks. These approaches first compute the \textit{importance weight} of each parameter in the network during (or after) learning a particular task. Then, while learning subsequent tasks, changes to the \textit{important} parameters are avoided or penalized. All regularization-based approaches, such as Elastic Weight Consolidation (EWC) \cite{kirkpatrick2017overcoming}, online EWC \cite{schwarz2018progress}, and Synaptic Intelligence \cite{zenke2017continual}, follow this principle, but differ in how importance weights are computed.
\textit{Replay-based approaches} either i) store original data from the previous tasks, replay them with the data from the current task while optimizing network parameters jointly over all data, or ii) use a generative model to generate pseudo-data to be replayed with the real data samples. Incremental Classifier and Representation Learning (iCaRL)  \cite{rebuffi2017icarl}, Deep Generative Replay (DGR) \cite{shin2017continual}, Deep Generative Replay with Distillation \cite{vandeven2018generative,vandeven2019three}, and Random Path Selection (RPS-net) \cite{rajasegaran2019random} are examples of replay-based approaches.

Regularization based CL approaches are useful as they neither store data from the previous tasks nor add more layers or nodes to the network with each incoming task, and thus avoid data storage and architectural complexity issues. However, with a fixed capacity and no access to previous data -- not even generated data -- these approaches struggle for challenging datasets under domain and class-incremental settings. For simpler tasks, regularization based CL approaches perform well, and hence we consider their vulnerability for such tasks in the experiments and results section.

Replay based approaches are more successful in learning data from continuously changing distributions, therefore we explore the vulnerability of these approaches for various datasets and scenarios in greater detail. However, it is important to reemphasize that replay based approaches that replay original data from previous tasks violate the basic assumption of incremental learning that data from previous tasks are not available. Furthermore, for some applications, such as in medical domain, there is also a privacy risk in storing and replaying original data \cite{de2019continual}. Therefore, we consider generative approaches that learn the data distribution over  previous tasks and then generate pseudo-samples from the previous tasks during the training on the current task. More specifically, we consider EWC, online EWC, and SI as examples of the regularization-based approaches, with DGR and DGR with distillation as examples of generative replay-based approaches as they do not require to store original data.

We suspect that the ability of these approaches to learn incrementally by adapting to new data makes them particularly vulnerable to adversarial attacks. We show that a determined adversary can easily force any of these approaches to learn small amount of misinformation represented with a backdoor tag during sequential training of different tasks, even if such backdoor tags are visually imperceptible to humans. This imperceptible misinformation can then be used for targeted misclassification of the samples of any prior task to the adversary's desired false label at inference time. The idea of presenting imperceptible information at test time is similar in spirit to the targeted evasion attacks using adversarial examples \cite{goodfellow2014explaining}, but our attack procedure is entirely different: here, we are not assuming a white-box scenario where the attacker knows the model parameters, nor are we \textit{generating} imperceptible noise to be added to the test images, rather we are forcing the model in a black-box setting to learn an imperceptible information during training of any single task. 

In preliminary stages of this work, we discussed vulnerability of importance based domain adaptation to poisoning attacks in \cite{umer2018adversarial,umer2019vulnerability}, the vulnerability of EWC under task incremental setting to \textit{perceptible} backdoor attacks in \cite{Umer2020TargetedFA},
and the vulnerabilities of regularization and generative replay approaches considering only the simple MNIST dataset and where the attack was launched on all tasks in \cite{umer2021adversarial}. This effort significantly expands the prior work to not only the more challenging SVHN and CIFAR10 datasets, but also to a more realistic scenario where the attacker inserts imperceptible misinformation to only a single task (as opposed to all tasks in prior work) of its own choosing. Manipulating the test time performance of the target task through misinformation only provided at the single time step represents a more realistic and also a more challenging scenario for an attacker.

\section{Adversarial Machine Learning (AML)}
AML explores the vulnerabilities of machine learning algorithms to malicious attacks. Two major types of such attacks are \cite{huang2011adversarial}: i) \textit{causative} (or \textit{poisoning}) attacks, which add strategically-chosen malicious data points into the training data \cite{umer2019vulnerability, biggio2012poisoning, umer2018adversarial} to impair future generalization capabilities of the classifier; and ii) \textit{exploratory} (or \textit{evasion}) attacks, which exploit misclassification at the test time to discover a model's blind spots or to extract information about the data or the model itself \cite{biggio2013evasion}. Since CL involves retraining / updating the model with each new data, the adversary's choice in targeting CL algorithms is typically a poisoning attack \cite{Umer2020TargetedFA, umer2021adversarial}.

Backdoor attacks are a specific class of poisoning attacks \cite{gu2017badnets,shafahi2018poison}, typically launched against ANNs in computer vision applications. In backdoor attacks that we consider here, the malicious samples are created by tagging a small portion of training images with a special \textit{backdoor pattern}. The adversary assigns a false label to these malicious samples, which are then added to the training set. The model is trained with both original as well as tagged and mislabeled images. The attacker's goal is to force the model to learn an association between the backdoor pattern and the false label. Once the model learns this association, it performs well on clean (untagged) test data during the testing (``inference'') stage, while causing targeted misclassification on data with the backdoor pattern. The attacker is then empowered to employ \textit{targeted evasion attacks} against the model by applying the backdoor tag to any test image of its choice. Since clean images are correctly classified, this attack is particularly difficult to detect using standard defenses; an unsuspecting victim may only slowly (or never) become aware of the vulnerability in their model, as the model performs largely as expected \textit{except} in the presence of the backdoor tag.

\textbf{Attacking Continual Learners:} In this work, we explore the impact of the backdoor attacks in the context of continual learning. Our major contributions in this work are as follows.

First, we examine the vulnerability of several state of the art CL models to the \textit{gradual} presentation of small amount of misinformation. Conventional backdoor attacks assume a large amount of tagged and mislabeled samples, with some also assuming access to both the data as well as the model parameters \cite{gu2017badnets, liu2017trojaning}. To demonstrate the critical nature of this vulnerability, we relax the conventional attack scenario that requires a large amount of tagged and mislabeled samples at once: we show that an attack can be very effective even with a small number of malicious samples, and only with access to data but \underline{not} model parameters. Such a strategy then allows the attacker to add small amount of misinformation at any time-step of its choosing, reducing the likelihood of detection.

Second, in conventional backdoor attacks the backdoor pattern is 
generally visible to naked eye, and thus can easily be detected through simple visualization if audited \cite{nguyen2021wanet}. To make the attack undetectable and further demonstrate the vulnerability of the model, our attack strategy uses a backdoor pattern that is imperceptible to humans. We should note that some recent works also try imperceptible patterns  \cite{li2020invisible, nguyen2021wanet}, but they are used in very different contexts (such as non-continual settings), and also use complicated procedures to generate the imperceptible pattern. We instead propose a computationally inexpensive, simple and effective strategy describe below. 

Finally, we show backdoor samples need not be inserted into the training data of the specific task being targeted. In fact, we demonstrate that the backdoor information can be inserted at \textit{any} attacker-chosen time step(s) \textit{in the future}, and still have significant impact on the model's performance on a \textit{previously-learned} task that was not attacked at the time of its training. This capability of launching an attack at any time is critical, as it makes detection by the defender even more difficult compared to a conventional backdoor attack. The conventional backdoor attacks are successful because the training and test data are assumed to be drawn from the same distribution. In CL setting, however, the tasks can -- and usually are -- different, meaning the data for different tasks have different distributions. Therefore, in such CL settings, providing backdoor pattern in the training data of the target task during training, and then using that same backdoor to attack the test data of the same target task is trivial, as it exactly represents the conventional backdoor strategy. In this effort, we demonstrate a far more challenging but sinister attack strategy, i.e., launching an attack to the target task at any time of its choosing during training of other unrelated task(s).

\subsection{Imperceptible Backdoor Pattern Generation}
\begin{figure}[!t]
\centering
\subfloat[]{\includegraphics[width=0.7in]{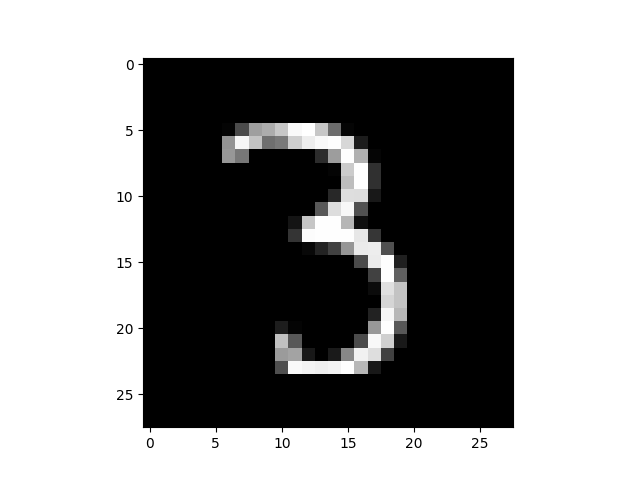}%
\label{img_org}}
\hfil
\subfloat[]{\includegraphics[width=0.81in]{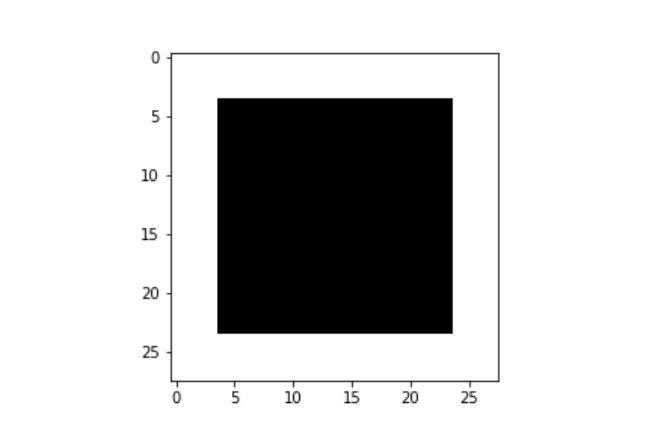}%
\label{img_vis_fr}}
\hfil
\subfloat[]{\includegraphics[width=0.7in]{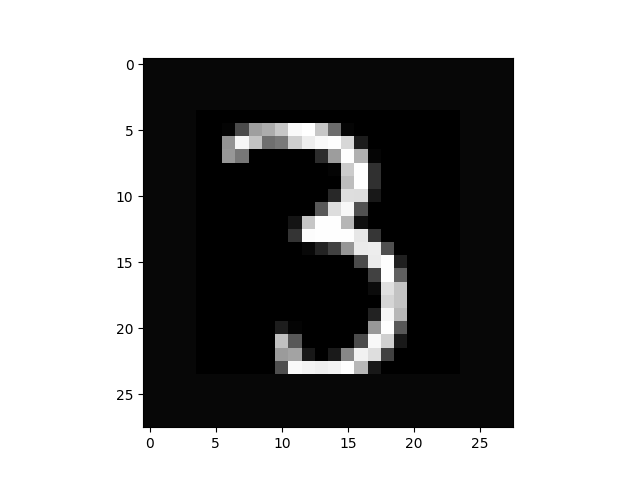}%
\label{img_inv_fr}}
\caption{Imperceptible backdoor pattern for MNIST images: (a) Original image; (b) backdoor tag as a frame; (c) image containing imperceptible frame as a backdoor pattern.}
\label{fig:mnist_inv_bd}
\end{figure}

Backdoor patterns are generally added by manipulating some of the lesser relevant features, such as modifying the intensity of edge / corner pixels, or adding a specific pattern or sticker onto these pixels. Here, we use a frame of pixels that are imperceptible to human eye. For MNIST gray scale images, for example, most pixels around the perimeter of the image are black (zero). Therefore, we use a rectangular frame around the perimeter and slightly increase associated pixel values from 0 to 0.03, a change that is completely imperceptible to the human eye. We use this frame as \textit{the} misinformation for MNIST variants in our experiments. Mathematically, let $x$ be the clean image and $r_f$ be an image equal in size to the clean image. All pixel in $r_f$ are zero except a rectangular frame of width equal to 5 pixels, where the pixel values are set to 0.03. The malicious image $x_m$ can be obtained by simple addition of these two images as $x_m = x + r_f$, as seen in Figure \ref{fig:mnist_inv_bd}.

\begin{figure}[!t]
\centering
\subfloat[]{\includegraphics[width=0.81in]{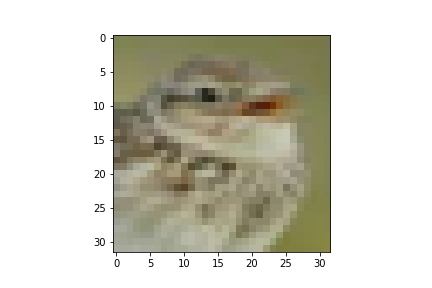}%
\label{img_org_cifar_f}}
\hfil
\subfloat[]{\includegraphics[width=0.81in]{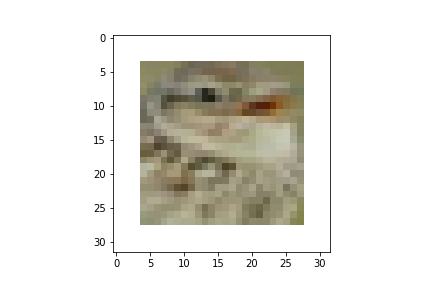}%
\label{img_vis_fr_cifar_f}}
\hfil
\subfloat[]{\includegraphics[width=0.81in]{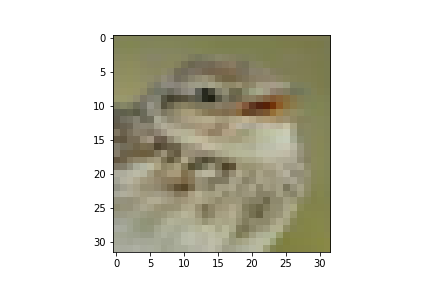}%
\label{img_inv_fr_wm_cifar_f}}
\hfil
\subfloat[]{\includegraphics[width=0.81in]{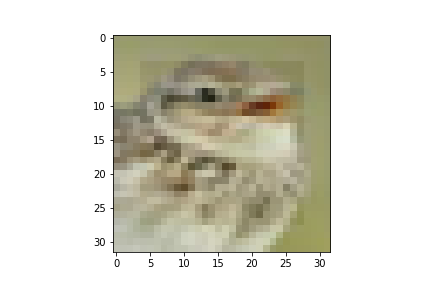}%
\label{img_vis_fr_wm_cifar_f}}
\caption{Attacking Cifar10 images: (a) original image; (b) framed image; (c) image containing \textit{imperceptible} frame as a backdoor pattern with $\epsilon$ of 0.01; (d) image containing \textit{perceptible} frame as a backdoor pattern with $\epsilon$ of 0.1.}
\label{fig:cifar_inv_bd_f}
\end{figure}

Generating an imperceptible backdoor pattern for realistic colored images requires a bit more care, as even slightly altered color pixels are very visible. To do so, we set $r_f$ to the original image with a frame whose values are now set to one, and use a \textit{watermark} as a backdoor pattern as the sum of the clean image $x$ and a weighted framed image $r_f$. The weight of $r_f$ is set to $\epsilon$ to obtain $x_{m} = x + \epsilon * r_f$. The imperceptibility of the backdoor pattern is controlled by $\epsilon$: smaller values make the pattern less noticeable to humans. A value of $\epsilon = 0.01$ results in a very imperceptible backdoor pattern. For Cifar10 dataset, a sample clean image, framed image, an attack image with an invisible backdoor pattern ($\epsilon=0.01$),  as well as a visible pattern  (with $\epsilon=0.1$) are shown in Figures \ref{img_org_cifar_f} , \ref{img_vis_fr_cifar_f}, \ref{img_inv_fr_wm_cifar_f}, and \ref{img_vis_fr_wm_cifar_f}, respectively.

\subsection{Attacking Regularization Approaches}
The regularization based CL model's generalized loss function $\ell(f_{\theta})$ for learning the current task $t$ can be written as: 
\begin{equation}
\label{reg_loss_eq}
    \ell(f_{\theta}) = \ell[f_{\theta}(X^t),y^t] + \lambda \sum\nolimits_i I_{t-1,i} (\theta_{t,i} - \theta^*_{t-1,i})^2
\end{equation}
where $(X^t, y^t)$ are the data-label pair and $\ell[f_{\theta}(X^t),y^t]$ is the model's loss on the current task $t$;  $I_{t-1,i}$ is the $i^{th}$ parameter's importance matrix computed for the previous task $t-1$; $\theta^*_{t-1,i}$ is the optimal value of the $i^{th}$ parameter learned for the previous task $t-1$; and $\lambda$ is the regularization coefficient.

Let $X^t_{b}$ represent the backdoor samples inserted into the training data of the current task and $y_{b}^t$ be their corresponding incorrect labels as chosen by the attacker. The loss function 
to be minimized during the training of current task is then:
\begin{equation}
\label{reg_loss_w_BD}
\begin{split}
    \ell(f_{\theta}) = &\ell[f_{\theta}(X^t \cup X^t_{b}),(y^t \cup y_{b}^t)] \\
                        &+ \lambda \sum\nolimits_i I_{t-1,i}(\theta_{t,i} - \theta^*_{t-1,i})^2
\end{split}
\end{equation}

Once the model is trained,
the test examples from the previous task are presented to the model. 
For those test examples that do not contain the backdoor pattern, i.e., $X_{test}^{t}$, the prediction of the model should be (mostly) correct, $f(X_{test}^{t}) \approx y_{test}^{t}$. Test samples that contain the backdoor pattern, however, will be misclassified as the attacker's desired label, resulting in $f(X_{test_ b}^{t}) = y_{test_b}^{t} \neq y_{test}^{t}$.

\textbf{Attacker's Approach:}
Regularization based approaches typically perform well on those CL datasets where the data distributions among tasks are different but related under domain incremental learning (DIL), however, they fail under the more challenging class incremental learning scenarios (CIL) even for simple MNIST variants \cite{vandeven2019three}. This impediment can be attributed to their fixed capacity, with no memory of past experiences. As the number of unrelated tasks grow, the fixed capacity model without any memory is unable to avoid the bias caused by unrelated and different distributions. Therefore, to attack these approaches, the attacker can simply leverage the \textit{relatedness} of distributions and insert a fixed imperceptible backdoor pattern into only a small amount, e.g. 1\%, of the training data of task(s) other than the attacker's targeted task, as \textit{misinformation}. Furthermore, as we demonstrate, since the output label space is fixed in DIL for every task, it is possible to degrade the test performance on the target task with the imperceptible backdoor pattern inserted into the training data of \textit{unrelated non-targeted} tasks, as a form of misdirection.

\subsection{Attacking Generative Replay Based Approaches}
Generative replay approaches, such as \textit{deep generative replay} (DGR) \cite{shin2017continual}, and \textit{DGR with distillation} (DGR-D)\cite{vandeven2018generative, vandeven2019three} use a variational autoencoder (VAE) \cite{kingma2013auto} to generate representative pseudo-samples of the previous tasks, and "reuse" those samples for reminding the model to what was previously learned. DGR then minimizes the following loss function
\begin{equation}
\label{eq_dgr}
\begin{split}
    \ell(f_{\theta}) = & (1/T) \ell_{current}[f_{\theta}(X^t),y^t] + \\ &  (1-(1/T))\ell_{replay}[f_{\theta}(X^{t-1} \cup X^{t-2} \cup \dots X^1), \\ & \hspace{1.1in} y^{t-1} \cup y^{t-2} \cup \dots y^1]
\end{split}
\end{equation}
where $T$ is the number of tasks seen so far,  $\ell_{current}[f_{\theta}(X^t),y^t]$ is the loss on current data, and $\ell_{replay}[f_{\theta}(X^{t-1} \cup X^{t-2} \cup \dots X^1), y^{t-1} \cup y^{t-2} \cup \dots y^1]$ is the loss on the data replayed from all previous tasks. Here, $X^k$, $k = t-1,t-2, \dots, 1$ are the pseudo-generated samples provided by the VAE for all previous tasks and $y^k$ are their corresponding correct labels. In DGR-D, VAE also generates pseudo-samples from the previous task(s) to be replayed with the current task's training data, but instead of labeling the replay samples with the most likely category (hard targets), it labels these samples with the soft targets, i.e., the predicted probabilities for all classes obtained from the model after finishing training on the most recent task. The overall loss function minimized by DGR-D is then given as
\begin{equation}
\label{eq_dgr_w_dist}
\begin{split}
    \ell(f_{\theta}) = & (1/T) \ell_{current}[f_{\theta}(X^t),y^t] + \\ & (1-\frac{1}{T})\ell_{distillation}[f_{\theta}(X^{t-1} \cup X^{t-2} \cup \dots X^1), \\ &  \hspace{1.1in} (\Tilde{y}^{t-1} \cup \Tilde{y}^{t-2} \cup \dots \Tilde{y}^1)]
\end{split}
\end{equation}
where, $\Tilde{y}^k$, $k= t-1, t-2, ... , 1$ are the soft targets for the data from all previous tasks that are generated and replayed.

Again let $X^t_{b}$ represent the backdoor samples in training data of the current task and $y_{b}^t$ be their corresponding incorrect labels. The loss function for DGR to be minimized during the training of current task with the backdoor samples is then:
\begin{equation}
\label{eq_dgr_w_bd}
\begin{split}
    \ell(f_{\theta}) = & (1/T) \ell_{current}[f_{\theta}(X^t \cup X^t_{b}),(y^t \cup y_{b}^t)] + \\ & (1-(1/T))\ell_{replay}[f_{\theta}(X^{t-1} \cup X^{t-2} \cup \dots X^1), \\ & y^{t-1} \cup y^{t-2} \cup \dots y^1]
\end{split}
\end{equation}
whereas for DGR-D (DGR with distillation), we similarly have:
\begin{equation}
\label{eq_dgr_w_dist_w_bd}
\begin{split}
    \ell(f_{\theta}) = & (1/T) \ell_{current}[f_{\theta}(X^t \cup X^t_{b}),(y^t \cup y_{b}^t)] + \\ & (1-\frac{1}{T})\ell_{distillation}[f_{\theta}(X^{t-1} \cup X^{t-2} \cup \dots X^1), \\ & (\Tilde{y}^{t-1} \cup \Tilde{y}^{t-2} \cup \dots \Tilde{y}^1)]
\end{split}
\end{equation}

Upon training, the test examples $X_{test}^{t}$ that do not contain the backdoor pattern should have (mostly) correct classification at inference time, i.e.,  $f(X_{test}^{t}) \approx y_{test}^{t}$. Test samples with the backdoor pattern, however, will be misclassified as the attacker's desired label, resulting in $f(X_{test_b}^{t}) = y_{test_b}^{t} \neq y_{test}^{t}$.

\textbf{Attacker's Approach:} Generative replay  approaches can handle the more realistic and challenging class incremental learning scenarios (where both input and class distributions are different between tasks), as well as domain incremental learning cases (where only the marginal data distributions between tasks are different). The generative replay approaches accomplish this feat by replaying generated instances, which results in an expanding (non-fixed) capacity model.

For such a model, the attacker can shrewdly exploit the data replay feature, and force the model to learn the association between imperceptible backdoor pattern and the false label, even with very small amount of training data, but this time from the target task. As before, the attacker can use a misdirection and insert these \textit{target-task backdoor samples} during the training of the unrelated \textit{non-target tasks}. More specifically, since a generative replay approach replays pseudo-samples from prior tasks (including the target task that the attacker wants to attack at test time), the attacker can insert (or hide) a small number of samples (say, 1\%) from the training data of the target task with imperceptible backdoor pattern, into the training data of the current (non-target) task of its choosing. Once the misinformation is learned, it can be effectively used to cause targeted forgetting of the desired task at test time.

\section{Experiments \& Results}
We evaluate the vulnerability of continual learning (CL) algorithms against our attack models under several datasets commonly used in CL settings. 
First, we consider the two common CL variants of MNIST dataset, i.e, \textit{rotation} MNIST (R-MNIST) \cite{lopez2017gradient} and \textit{split} MNIST (S-MNIST) \cite{zenke2017continual}. For R-MNIST, we create a sequence of 5 tasks, each associated with a different randomly generated rotation in the interval $(0, \pi/3]$. Each task in R-MNIST is a 10-class problem: the first task is to classify original ten MNIST digits; each other task involves classifying the same ten digits with a different random rotation applied to all digits. S-MNIST also involves 5 tasks, where each task is a binary classification problem: the first task is to distinguish between digits 0 and 1, the second task is to distinguish between digits 2 and 3, and so on.

We then consider the more challenging split SVHN \cite{netzer2011reading} and split Cifar10 \cite{krizhevsky2009learning} datasets. SVHN dataset includes natural scene images obtained from house numbers in Google Street View images..
Cifar10 dataset involves images from ten categories (airplanes, cars, birds, cats, deer, dogs, frogs, horses, ships, and trucks). The SVHN and Cifar10 datasets are also split into five tasks to obtain split SVHN and split Cifar10 CL versions, where each task is a binary classification problem.

The attacker first chooses a \textit{target} task it wants the model to forget (misclassify) at test time. The attacker may choose any of the tasks as its desired target task. Without any loss of generality, we assume that the target task is Task 1. To degrade the test time performance of the target task, the attacker can insert malicious samples into the training data of \textit{all} non-target tasks (i.e., all tasks except Task 1 as done in our prior work \cite{umer2021adversarial}. However, in order to make the attack even more difficult to detect, we now restrict the attacker to add small amount of \textit{imperceptible} misinformation \textit{only at the last task} to degrade the test time performance of the \textit{first task}. Here, we assume that some amount of time has passed since the target task was learned, and only now does the attacker decides to attack the model - either because it only now wishes to attack, or more likely, it wants to hide its goals. From the attacker's perspective, attacking the first task after all other tasks have already been learned is the most challenging scenario. 

\subsection{Attacking Domain Incremental Learning}

\subsubsection{Attacking R-MNIST}
To attack R-MNIST, we added imperceptible backdoor pattern to randomly picked 1\% of the training data and inserted them into the training data of the \textit{last task}. Recall that under DIL, each task of R-MNIST contains the same 10 classes, therefore, without any loss of generality, we pick the incorrect label the attacker wants to assign to malicious samples is of class 0. Hence, 1\% malicious data are randomly drawn from all classes except 0. The model is then trained with 10-class clean data for the first four tasks, and with the compromised data for the last task. After training on all five tasks, clean test examples are presented to the model at inference time for all tasks, except the targeted first task, for which the test examples (from all classes excluding class 0) included the imperceptible backdoor pattern.

The test accuracy on each task along with the average accuracy over all tasks are shown in Table \ref{table:accuracy_reg_rM_lt} and Table \ref{table:accuracy_gen_rM_lt} for each of the regularization (EWC, SI, and Online EWC), and the generative replay approaches (DGR and DGR-D), respectively; both under \textit{no} attack (NA) and attack scenarios.  All attack scenario results are presented with mean and standard deviations computed over ten independent runs.
Table \ref{table:accuracy_reg_rM_lt} and Table \ref{table:accuracy_gen_rM_lt} show that -- despite being presented only during the last task -- the merely 1\% malicious data with imperceptible backdoor pattern are sufficient to force all CL models to forget the first task, as the models' 88-95\% performance under no attack drops to about 10\% under the attack for Task 1. Note that the performances on non-targeted tasks are not affected, as intended.  Most importantly, the attacker is able to successfully force the model to misclassify all Task 1 test data (except class 0) to its \textit{intended and desired} target label (of class 0) achieving $\approx$10\% accuracy, regardless of the CL algorithm used.

\begin{table}[h]
\Large
\caption[]{Test accuracy of regularization approaches on rotation MNIST. 
Performance on Task 1 with attack is in bold.}
\label{table:accuracy_reg_rM_lt}
\centering
\begin{adjustbox}{width=0.49\textwidth}
\begin{tabular}{|m{1.1cm}|m{0.95cm}|m{2.5cm}|m{0.95cm}|m{2.5cm}|m{1.5cm}|m{2.5cm}|}
\hline
Tasks & EWC (NA) & EWC (With Attack) & SI (NA) & SI (With Attack) & Online EWC (NA) &  Online EWC (With Attack)
\\ \hline\vspace*{.02cm}
 T1 &  0.88 & \textbf{0.10 $\pm$ 2e-3}   &   0.88 &  \textbf{0.14 $\pm$ 5e-3} &  0.87 & \textbf{0.11 $\pm$ 2e-3}  \\ \hline\vspace*{.02cm}
T2 & 0.96 & 0.96 $\pm$ 3e-4   & 0.95 & 0.96 $\pm$ 8e-5 &  0.95 &  0.96 $\pm$ 1e-3\\ \hline\vspace*{.02cm}
T3 &  0.97 & 0.97 $\pm$ 3e-4   & 0.95 &  0.95 $\pm$ 8e-5 & 0.96 &  0.96 $\pm$ 5e-4 \\ \hline\vspace*{.02cm}
T4 & 0.97 & 0.97 $\pm$ 4e-4   &  0.94  &  0.94 $\pm$ 5e-5  & 0.96 &  0.96 $\pm$ 3e-4 \\ \hline\vspace*{.02cm}
T5 & 0.96 & 0.96 $\pm$ 5e-4   & 0.92 & 0.90 $\pm$ 1e-4 &   0.94 & 0.93 $\pm$ 8e-4 \\ \hline \hline\vspace*{.02cm}
Mean & 0.95 & 0.79 $\pm$ 7e-4   & 0.93 & 0.78 $\pm$ 1e-3 & 0.94 & 0.78 $\pm$ 1e-3\\ \hline
\end{tabular}
\end{adjustbox}
\end{table}

\begin{table}[h]
\footnotesize
\renewcommand{\arraystretch}{0.5}
\caption[]{Test accuracy of generative replay approaches on rotation MNIST. 
Performance on Task 1 with attack is in bold.}
\label{table:accuracy_gen_rM_lt}
\centering
\begin{adjustbox}{width=0.49\textwidth}
\begin{tabular}{|m{0.53cm}|m{0.7cm}|m{1.5cm}|m{0.95cm}|m{1.9cm}|}
\hline
Tasks & DGR (NA) & DGR (With Attack) & DGR-D (NA) & DGR-D (With Attack)
\\ \hline\vspace*{0.05cm}
 T1 & 0.95 & \textbf{0.11 $\pm$ 6e-3}   &  \hspace{0.1cm} 0.97 &  \textbf{0.11 $\pm$ 3e-3}   \\ \hline\vspace*{.02cm}
T2 & 0.97 & 0.97 $\pm$ 3e-4   & \hspace{0.1cm} 0.98 &  0.98 $\pm$ 3e-4 \\ \hline\vspace*{.05cm}
T3 & 0.97 & 0.98 $\pm$ 5e-4   &  \hspace{0.1cm} 0.98 &  0.98 $\pm$ 1e-4  \\ \hline\vspace*{.05cm}
T4 & 0.97 & 0.97 $\pm$ 3e-4   &\hspace{0.1cm} 0.98  &  0.98 $\pm$ 3e-4  \\ \hline\vspace*{.05cm}
T5 & 0.97 & 0.97 $\pm$ 9e-4   &  \hspace{0.1cm} 0.98 & 0.98 $\pm$ 2e-4  \\ \hline \hline\vspace*{.05cm}
Mean &  0.97 & 0.80 $\pm$ 2e-3  & \hspace{0.1cm} 0.98 &  0.81 $\pm$ 8e-4 \\ \hline
\end{tabular}
\end{adjustbox}
\end{table}

\subsubsection{Attacking S-MNIST} 

As before, the adversary adds 1\% malicious samples containing the imperceptible backdoor pattern into the training data of only the last task, and the incorrect label the attacker wants to assign to malicious samples is of class 0; note that in S-MNIST, however, each task for DIL scenario has two possible labels of 0 and 1. 

The model is trained with the clean training data for the first four tasks and compromised data on the last task. At test time, clean examples are presented to the model for all non-target tasks. For the target task, clean samples from class 0 and malicious samples from class 1 are presented. The test accuracy on individual tasks along with the average accuracy over all tasks are shown in Table \ref{table:accuracy_gen_sM_lt} for DGR, and DGR-D both with and without attack. As described earlier, regularization approaches are not included in this experiment, as they failed to perform well even under no attack scenario.

Table \ref{table:accuracy_gen_sM_lt} shows that the attacker assumes complete control of the CL model's memory of the first task, even with only 1\% of misinformation provided in the last task. Specifically, there is a significant drop in Task 1 performance from 96\% under no-attack to 45\% (essentially random guess on this two-class-per-task problem) when under attack, whereas non-target task performances experience no such similar drop.

\begin{table}[h]
\footnotesize
\renewcommand{\arraystretch}{0.5}
\caption[]{Test accuracy of generative replay approaches on split MNIST. 
Performance on Task 1 with attack is in bold.}
\label{table:accuracy_gen_sM_lt}
\begin{adjustbox}{width=0.49\textwidth}
\begin{tabular}{|m{0.53cm}|m{0.7cm}|m{1.5cm}|m{0.95cm}|m{1.7cm}|}
\hline
Tasks & DGR (NA) & DGR (With Attack) & DGR-D (NA) & DGR-D (With Attack)
\\ \hline\vspace*{.02cm}
 T1 &  0.96 & \textbf{0.46 $\pm$ 2e-3}   &  \hspace{0.1cm} 0.98 & \textbf{0.45 $\pm$ 2e-3}   \\ \hline\vspace*{.02cm}
T2 &  0.95 & 0.96 $\pm$ 3e-3   & \hspace{0.1cm} 0.98 &  0.98 $\pm$ 2e-3 \\ \hline\vspace*{.02cm}
T3 & 0.89 & 0.91 $\pm$ 6e-3   &  \hspace{0.1cm} 0.95 &  0.96 $\pm$ 2e-3  \\ \hline\vspace*{.02cm}
T4 & 0.98 & 0.98 $\pm$ 7e-4   &  \hspace{0.1cm} 0.98  & 0.99 $\pm$ 7e-4 \\ \hline\vspace*{.02cm}
T5 & 0.98 & 0.97 $\pm$ 3e-3   & \hspace{0.1cm} 0.98 &  0.97 $\pm$ 2e-3  \\ \hline \hline\vspace*{.02cm}
Mean & 0.95 & 0.86 $\pm$ 3e-3  &  \hspace{0.1cm} 0.97 & 0.87 $\pm$ 2e-3 \\ \hline
\end{tabular}
\end{adjustbox}
\end{table}

\subsubsection{Attacking S-SVHN}
We attack generative replay based approaches with the proposed attack scheme for S-SVHN dataset to determine the robustness of these approaches. Following the same procedure, 1\% malicious data are provided only during the training data of the last task. At inference time, the imperceptible misinformation are used for targeted misclassification of all examples of class 1 from the first task. The results for DGR and DGR-D are shown in Table \ref{table:accuracy_gen_sSV_lt}, which shows that generative replay is still immensely vulnerable to the tiny amount of misinformation presented during training only at the last time step. Just to ensure that the attacker is achieving its intended goal of causing \textit{targeted} misclassification of all class 1 examples of the target task, we also computed the false negative rate or attack success rate, which were 98\% for DGR and 100\% for DGR-D.

\begin{table}[h]
\footnotesize
\renewcommand{\arraystretch}{0.5}
\caption[]{Test accuracy of generative replay approaches on split SVHN. Performance on Task 1 with attack is in bold.}
\label{table:accuracy_gen_sSV_lt}
\centering
\begin{adjustbox}{width=0.49\textwidth}
\begin{tabular}{|m{0.53cm}|m{0.7cm}|m{1.5cm}|m{0.95cm}|m{1.7cm}|}
\hline
Tasks & DGR (NA) & DGR (With Attack) & DGR-D (NA) & DGR-D (With Attack) 
\\ \hline\vspace*{.02cm}
 T1 & 0.63 & \textbf{0.14 $\pm$ 0.01}  &  \hspace{0.1cm} 0.55 & \textbf{0.15 $\pm$ 0.01}   \\ \hline\vspace*{.02cm}
T2 & 0.52 & 0.48 $\pm$ 0.03   &  \hspace{0.1cm} 0.59 &  0.57 $\pm$ 0.03 \\ \hline\vspace*{.02cm}
T3 &  0.80 & 0.76 $\pm$ 0.02   & \hspace{0.1cm} 0.84 &  0.84 $\pm$ 0.02  \\ \hline\vspace*{.02cm}
T4 &  0.95 & 0.93 $\pm$ 0.01   & \hspace{0.1cm} 0.96  & 0.95 $\pm$ 4e-3 \\ \hline\vspace*{.02cm}
T5 &  0.93 & 0.92 $\pm$ 0.01   &  \hspace{0.1cm} 0.94 &  0.94 $\pm$ 0.01  \\ \hline \hline\vspace*{.02cm}
Mean & 0.77 & 0.65 $\pm$ 0.01  & \hspace{0.1cm} 0.78 & 0.69 $\pm$ 0.01 \\ \hline
\end{tabular}
\end{adjustbox}
\end{table}

\subsubsection{Attacking S-Cifar10}
As before, the misinformation (1\% falsely labeled malicious samples containing imperceptible backdoor pattern) is only provided in the training data of the last task.
The results are summarized in Table \ref{table:accuracy_gen_sC_lt}, which again shows that the attacker can force a significant drop in Task 1 test performance, while other tasks remain unaffected. The false negative rate for DGR was 99\% and for DGR-D was 100\%, which shows that the attacker was able to force the model to misclassify its intended class instances.

\vspace{-0.1in}
\begin{table}[h]
\footnotesize
\renewcommand{\arraystretch}{0.5}
\caption[]{Test accuracy of generative replay approaches on split Cifar10. 
Performance on Task 1 with attack is in bold.}
\label{table:accuracy_gen_sC_lt}
\centering
\begin{adjustbox}{width=0.49\textwidth}
\begin{tabular}{|m{0.53cm}|m{0.7cm}|m{1.5cm}|m{0.95cm}|m{1.7cm}|}
\hline
Tasks & DGR (NA) & DGR (With Attack) & DGR-D (NA) & DGR-D (With Attack) 
\\ \hline\vspace*{.02cm}
 T1 &  0.81 & \textbf{0.40 $\pm$ 0.03}   &  \hspace{0.1cm} 0.80 & \textbf{0.34 $\pm$ 0.03}   \\ \hline\vspace*{.02cm}
T2 & 0.64 & 0.61 $\pm$ 0.02   &  \hspace{0.1cm} 0.63 &  0.63 $\pm$ 0.01 \\ \hline\vspace*{.02cm}
T3 &  0.69 & 0.68 $\pm$ 0.02  & \hspace{0.1cm} 0.60 &  0.62 $\pm$ 0.03  \\ \hline\vspace*{.02cm}
T4 & 0.69 & 0.72 $\pm$ 0.02   &  \hspace{0.1cm} 0.62  & 0.65 $\pm$ 0.04 \\ \hline\vspace*{.02cm}
T5 & 0.95 & 0.92 $\pm$ 1e-3   &  \hspace{0.1cm} 0.93 &  0.93 $\pm$ 0.02  \\ \hline \hline\vspace*{.02cm}
Mean &  0.75 & 0.67 $\pm$ 0.02  &  \hspace{0.1cm} 0.72 & 0.63 $\pm$ 0.03 \\ \hline
\end{tabular}
\end{adjustbox}
\end{table}

It is important to emphasize here that for all these cases, the proposed attack is not only acute but also more harder to detect as the attack samples are provided only during the last task, and the attacker is able to force the misclassification not just to any class label, but its intended class label, allowing this attack strategy to serve as both a poisoning and evasion attack. 
In other words, these results show utmost vulnerability of CL algorithm to the proposed attack vector.

\subsection{Attacking Class Incremental Learning (CIL)}

\subsubsection{Attacking R-MNIST}
Recall that CIL refers to subsequently learning new classes. The model is trained with clean data for the first four tasks followed by the last task with compromised data, with each task introducing ten new classes for R-MNIST. The model is then presented with backdoor tagged test samples for the target task and clean samples for non-target tasks. The attacker inserts 1\% malicious samples with imperceptible misinformation to the \textit{last task's} training data only. Target task test samples included the same imperceptible backdoor pattern in all classes, except class 0, the attacker's choice for the incorrect label for all other classes. The results are shown in Table \ref{table:accuracy_gen_rM_CIL_lt} for both DGR and DGR-D algorithms. As in other cases, we see that the attack causes significant drop from 73\% to 8\% -- below 10\%, random chance for a 10 class problem -- for the target task, with no such drop for non-target tasks, indicating the effectiveness of the proposed attack mechanism on generative replay approaches.

\begin{table}[h]
\footnotesize
\renewcommand{\arraystretch}{0.5}
\caption[]{Test accuracy of generative replay approaches on rotation MNIST under CIL setting.} 
\label{table:accuracy_gen_rM_CIL_lt}
\centering
\begin{adjustbox}{width=0.49\textwidth}
\begin{tabular}{|m{0.53cm}|m{0.7cm}|m{1.5cm}|m{0.95cm}|m{1.7cm}|}
\hline
Tasks & DGR (NA) & DGR (With Attack) & DGR-D (NA) & DGR-D (With Attack)  
\\ \hline\vspace*{.02cm}
 T1 & 0.73 & \textbf{0.08 $\pm$ 0.01}   &  \hspace{0.1cm} 0.80 & \textbf{0.10 $\pm$ 3e-3}   \\ \hline\vspace*{.02cm}
T2 & 0.83 & 0.87 $\pm$ 0.02  &  \hspace{0.1cm} 0.94 &  0.94 $\pm$ 2e-3 \\ \hline\vspace*{.02cm}
T3 &  0.87 & 0.82 $\pm$ 0.03  &  \hspace{0.1cm} 0.93 &  0.93 $\pm$ 4e-3  \\ \hline\vspace*{.02cm}
T4 & 0.93 & 0.93 $\pm$ 0.01   &  \hspace{0.1cm} 0.94  & 0.95 $\pm$ 1e-3 \\ \hline\vspace*{.02cm}
T5 &  0.97 & 0.97 $\pm$ 2e-3   & \hspace{0.1cm} 0.97 &  0.98 $\pm$ 4e-4  \\ \hline \hline\vspace*{.02cm}
Mean &  0.87 & 0.73 $\pm$ 0.01  & \hspace{0.1cm} 0.92 & 0.78 $\pm$ 2e-3 \\ \hline
\end{tabular}
\end{adjustbox}
\end{table}

\subsubsection{Attacking S-MNIST}
Split MNIST naturally favors class incremental learning as it involves five different sequential binary problems. The goal of the continual learning model is to correctly classify all ten classes after it is trained on mutually exclusive subsets of classes, two class at a time. 

The same attack strategy is used as discussed above, and the results are shown in Table \ref{table:accuracy_gen_sM_CIL_lt}. Once again, we observe that the attacker can easily create a false memory at test time, causing targeted misclassification of all samples of the target task that contain the imperceptible backdoor pattern to the attacker's incorrect label of choice (label 0 in our case) even with the small amount of misinformation provided in the last task only. We note that -- while successful in tackling challenging CIL scenarios for simple gray scale MNIST based continual variants under no attack -- generative replay based approaches also fail to handle more complex continual datasets such as split SVHN and split CIFAR10 even when there is no attack \cite{shen2020generative,lesort2019generative}, making it unnecessary to attack them for split CIFAR10 and split SVHN datasets.

\begin{table}[h]
\footnotesize
\renewcommand{\arraystretch}{0.5}
\caption[]{Test accuracy of generative replay approaches on Split MNIST under CIL setting.} 
\label{table:accuracy_gen_sM_CIL_lt}
\centering
\begin{adjustbox}{width=0.49\textwidth}
\begin{tabular}{|m{0.53cm}|m{0.7cm}|m{1.5cm}|m{0.95cm}|m{1.7cm}|}
\hline
Tasks & DGR (NA) & DGR (With Attack) & DGR-D (NA) & DGR-D (With Attack)  
\\ \hline\vspace*{.02cm}
 T1 &  0.84 & \textbf{0.43 $\pm$ 4e-3}   &  \hspace{0.1cm} 0.95 & \textbf{0.44 $\pm$ 3e-3}   \\ \hline\vspace*{.02cm}
T2 &  0.87 & 0.87 $\pm$ 5e-3   & \hspace{0.1cm} 0.90 &  0.89 $\pm$ 3e-3 \\ \hline\vspace*{.02cm}
T3 & 0.88 & 0.87 $\pm$ 0.01  & \hspace{0.1cm} 0.89 &  0.89 $\pm$ 0.01  \\ \hline\vspace*{.02cm}
T4 & 0.95 & 0.93 $\pm$ 3e-3 & \hspace{0.1cm} 0.96  & 0.96 $\pm$ 2e-3 \\ \hline\vspace*{.02cm}
T5 & 0.97 & 0.97 $\pm$ 2e-3   &  \hspace{0.1cm} 0.96 &  0.97 $\pm$ 2e-3  \\ \hline \hline\vspace*{.02cm}
Mean &  0.90 & 0.81 $\pm$ 5e-4  &  \hspace{0.1cm} 0.93 & 0.83 $\pm$ 4e-3 \\ \hline
\end{tabular}
\end{adjustbox}
\end{table}

\section{Conclusion \& Future Work}
We have shown that both of the two realistic and important scenarios of continual learning, namely domain and class incremental learning, are vulnerable to backdoor poisoning attacks. We show that an attacker can take advantage of a CL algorithm's very ability to continuously learn new information over time, and use that ability against itself by forcing to retain small amount of \textit{misinformation}. We hence showed that, just like human brain can be forced to learn misinformation while learning sequentially, so can ANN models when tasked to learn under continual learning settings. More importantly, we showed that an adversary can implant such misinformation into the model's knowledge and force it to misclassify all classes into its label of choice even by using a very small amount of malicious training data and even by attacking an unrelated future task. These experiments hence demonstrate the extreme vulnerability of CL approaches to such attacks under both domain and class incremental settings. 

Our natural future work is to develop an appropriate defensive solution against such attacks. This work accentuates the critical need for any future algorithm seeking artificial general intelligence to be robust against adversarial threats under continual / incremental learning settings. It is of paramount importance to prioritize the development of robust continual learning algorithms, so that they can learn from streaming data while remaining secure against deliberate misdirection.






%
\bibliographystyle{IEEEtran}
%

\bibliography{bibliography}{}

%
%

%
\vspace{-0.39in}
\begin{IEEEbiographynophoto}{Muhammad Umer}
is a Ph.D. student in Electrical and Computer Engineering at Rowan University. His current area of interest is in exploring vulnerabilities and defenses of continual learning algorithms against adversarial attacks. 
\end{IEEEbiographynophoto}

\begin{IEEEbiographynophoto}{Robi Polikar} is Professor and Department Head of Electrical and Computer Engineering at Rowan University. His prior work has focused on ensemble based approaches for incremental learning, and learning in non stationary environments. His current area of interest is in adversarial machine learning.

\end{IEEEbiographynophoto}






\end{document}